%% file: main.tex
\documentclass{l4dc2025}

\input{preamble/preamble}
\graphicspath{{./figures/}}

\title[Failure Probability Estimation using State-Dependent Proposals]{Failure Probability Estimation for Black-Box Autonomous Systems using State-Dependent Importance Sampling Proposals}
\usepackage{times}

\author{\Name{Harrison Delecki} \Email{hdelecki@stanford.edu}\\
\Name{Sydney M. Katz} \Email{smkatz@stanford.edu}\\
\Name{Mykel J. Kochenderfer} \Email{mykel@stanford.edu}\\
\addr Department of Aeronautics and Astronautics\\
Stanford University, Stanford, CA 94305, USA}

\begin{document}

\maketitle

\begin{abstract}%
Estimating the probability of failure is a critical step in developing safety-critical autonomous systems. Direct estimation methods such as Monte Carlo sampling are often impractical due to the rarity of failures in these systems. Existing importance sampling approaches do not scale to sequential decision-making systems with large state spaces and long horizons. We propose an adaptive importance sampling algorithm to address these limitations. Our method minimizes the forward Kullback–Leibler divergence between a state-dependent proposal distribution and a relaxed form of the optimal importance sampling distribution. Our method uses Markov score ascent methods to estimate this objective. We evaluate our approach on four sequential systems and show that it provides more accurate failure probability estimates than baseline Monte Carlo and importance sampling techniques. This work is open sourced.\footnote[1]{\href{https://github.com/sisl/SPAIS.jl}{https://github.com/sisl/SPAIS.jl}}

\end{abstract}


\section{Introduction}
Autonomous systems are increasingly being considered for safety-critical domains such as aviation \citep{kochenderfer2012next, owen2019acasxu}, autonomous driving \citep{badue2021selfdrivingsurvey}, and home robotics \citep{zachiotis2018survey}. These autonomous systems operate sequentially by taking actions after receiving observations of the environment. For example, an autonomous vehicle must continuously process sensor data, predict the movements of other vehicles, and make decisions about steering and braking. The safety of these systems must be rigorously validated in simulation before deployment. One way to quantify system safety is to estimate the probability of system failure such as a collision of an autonomous vehicle with other agents. Estimating the probability of failure for autonomous systems may highlight weaknesses and uncover potentially dangerous scenarios. 

There are three key challenges associated with estimating the probability of failure in sequential autonomous systems. First, failures tend to be rare when the system is designed to be safe. Estimating the probability of failure using simple methods like Monte Carlo sampling may require an enormous number of expensive simulations. Second, the search space over failure events is very high-dimensional because autonomous systems operate over large state spaces and long time horizons. Third, autonomous systems can exhibit multimodal failures, requiring validation techniques that can adequately capture diverse behaviors.

Many previous approaches have attempted to address these challenges through importance sampling techniques. Importance sampling techniques aim to sample from alternative distributions that assign higher likelihood to failure events \citep{owen2000safe}. These methods often struggle to scale to high-dimensional, long-horizon problems because they rely on simple parametric proposals that are difficult to design.  Extensions of importance sampling to sequential problems using dynamic programming and reinforcement learning show promise but remain limited to small discrete state spaces and short time horizons \citep{Chryssanthacopoulos2010, corso2022deep}.

\begin{figure}[!t]
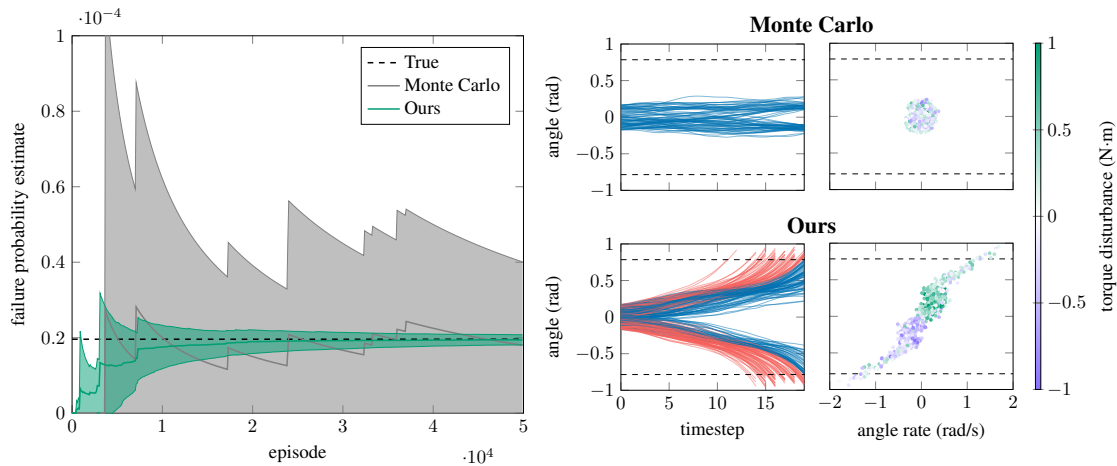

    \centering
    \includestandalone[width=1.0\textwidth]{showcase/showcase_standalone}
    \caption{Comparison of the proposed approach and Monte Carlo (MC) sampling for estimating failure probabilities in an inverted pendulum with a rule-based controller under torque disturbances. Failures occur when the magnitude of the pendulum's angle from the vertical exceeds a threshold. The left figure shows the estimated failure probability using our method during training compared to MC. Our method is more accurate with fewer samples than MC. The middle two figures compare $200$ pendulum angle trajectories over time using samples from MC and from ours after training. Our method discovers failure trajectories (red) across  modes, while all MC samples are safe (blue). The two right figures show torque disturbances applied at each state in the sampled trajectories. Our method learns to add positive torque disturbances for positive angles (left from vertical) and vice versa, enabling efficient sampling for failure probability estimation.}
    \label{fig:showcase}
\end{figure}

In this work, we aim to enable efficient failure probability estimation for black-box sequential systems through importance sampling. Inspired by previous work, we decompose the high-dimensional sampling problem over system trajectories using a sequential state-dependent proposal distribution. We optimize the proposal by minimizing the forward Kullback–Leibler (KL) divergence between the sequential proposal and an approximate distribution over failure trajectories. We estimate gradients of the objective  using a Markov score ascent method \citep{naesseth2020msc, kim2022msa}, which approximates samples from the target distribution using MCMC. In addition to optimizing the proposal, the algorithm uses these samples to compute an importance sampling estimate of the probability of failure. The impact of our approach is illustrated in \cref{fig:showcase}. In summary, our contributions are as follows:
\begin{itemize}
    \itemsep0em 
    \vspace{-1mm}
    \item We propose a sequential state-based importance sampling proposal for failure probability estimation of autonomous systems.
    \item We optimize the proposal by minimizing the forward KL divergence to a tractable approximation of the optimal importance sampling distribution.
    \item We demonstrate that the proposed approach achieves more accurate estimates of the probability of failure compared to baselines on four sample problems.
\end{itemize}

\section{Related Work}
In this section, we first review broad categories of safety validation approaches. Then, we review failure probability estimation methods, including adaptive importance sampling techniques.

\paragraph{White-Box vs. Black-Box} Traditional validation methods take a white-box approach that uses internal information about the system under test to prove safety properties \citep{Schumann2001, Clarke2018}. In this work, we take a black-box approach. Black-box methods generally scale to large, complex systems because they do not require an internal model, but only pass inputs to the system and observe outputs. See the survey by \citet{Corso2021survey} for more on black-box methods. 

\paragraph{Falsification} A common approach to black-box safety validation of sequential systems is falsification, which aims to find individual failure trajectories. Falsification approaches use optimization \citep{deshmukh2017testing}, trajectory planning \citep{tuncali2019rapidly}, and reinforcement learning \citep{Akazaki2018falsification} to find specific inputs that lead to failure. Adaptive Stress Testing further incorporates trajectory likelihood to search for the most-likely failure event \citep{lee2020adaptive}. The drawback of falsification methods is that they tend to converge on a single failure, such as the most-likely or most severe, and do not explore the space of potential failures.

\paragraph{Failure Probability Estimation} Failure probability estimation provides a more comprehensive assessment of system safety by characterizing the distribution over failures. Some previous approaches rely on Markov chain Monte Carlo (MCMC) methods to sample failure events \citep{sinha2020neural, norden2019efficient}. These approaches are generally only feasible in low-dimensional problems, may require domain knowledge to perform well, or may require the system under test to be differentiable to scale to larger problems.  Adaptive Importance Sampling (AIS) methods iteratively improve a proposal distribution using samples from the proposal \citep{bugallo2017adaptive, rubinstein2004cross, cappe2004population}. The drawback of these methods is that they tend to become intractable in problems with long time horizons, and they are typically limited to modeling a small set of input parameters \citep{uesato2019rigorous}. Importance sampling algorithms have been extended to sequential problems by incorporating dynamic programming, but this approach is limited to small discrete state spaces \citep{Chryssanthacopoulos2010}. Finally, some work builds on reinforcement learning by iteratively improving a proposal using policy gradient methods \citep{corso2022deep}. Due to the large variance of the estimator, this approach does not scale to long horizons or large action spaces. In this work, we propose an AIS algorithm for sequential problems with continuous state spaces and long horizons.

\section{Background}
In this section, we provide necessary background on safety validation for sequential autonomous systems, failure probability estimation, and importance sampling.

\subsection{Safety Validation for Sequential Systems}

Consider a system under test that takes actions $a$ in an environment after receiving observations $o$ of its state $s$. We assume states may transition stochastically according to a transition model $p(s' \mid s, a)$ and state observations may depend on imperfect sensors modeled as $p(o \mid s)$. The system under test may also act stochastically depending on the observation with policy $\pi(a \mid o)$.

In safety validation, we search for failure events by controlling sources of randomness in the state transitions, observations, and system under test. We use disturbances, denoted by $x$, to control sources of randomness. We assume disturbances are continuous values in the space $\mathcal{X} \subseteq \mathbb{R}^{D_x}$ where $D_x \in \mathbb{N}$ is the dimension of the disturbance space. Given a state $s_t$ and disturbance $x_t$ at time $t$, we can simulate the next timestep $t+1$ according to
\begin{equation}
    o_t \sim p(\cdot \mid s_t, x_t), \quad a_t \sim \pi(\cdot \mid o_t, x_t), \quad s_{t+1} \sim p(\cdot \mid s_t, a_t, x_t)
\end{equation}
where different components of the disturbance $x_t$ may be applied to the observation, system under test, and transition. Given a set of disturbances up to horizon $T$, we can simulate the system under test with disturbances to compute a system trajectory $\tau = (s_1, a_1, o_1, x_1, \ldots,s_T, a_T, o_T, x_T)$.

We assume that disturbances are sampled according to a distribution that may depend on the current system state $d(x \mid s)$. The disturbance distribution reflects how likely different disturbances are to occur in the real operating environment, and is typically designed using real-world data or expert knowledge. This formulation could also consider disturbances that depend on observations and actions, but we focus only on state-dependent disturbances for simplicity. Using the disturbance distribution, we can define a distribution over system trajectories
\begin{equation}
    \label{eq:nominal-density}
    p(\tau) = p(s_1) \prod_{t=1}^T p(s_{t+1} \mid s_t, a_t, x_t) p(o_t \mid s_t, x_t) \pi(a_t \mid o_t, x_t) d(x_t \mid s_t)
\end{equation}
where $p(s_1)$ is the initial state distribution. We denote this the nominal trajectory distribution since it represents the anticipated trajectory distribution in operation.


We evaluate the safety of system trajectories using an evaluation function $f(\tau) \in \mathbb{R}$. When $f(\tau)$ exceeds a failure threshold $\gamma$, we say that the trajectory is a failure event, with higher values indicating greater proximity to failure. We can express the distribution over all failure trajectories as the conditional distribution
\begin{equation}
    \label{eq:failure-distribution}
    p(\tau \mid \mathds{1} \{ f(\tau) \geq \gamma \}) = \frac{p(\tau) \mathds{1} \{ f(\tau) \geq \gamma \}}{\int p(\tau') \mathds{1} \{ f(\tau') \geq \gamma \} d\tau'}
\end{equation}
where $\mathds{1}\{\cdot\}$ is the indicator function and the denominator integrates over all trajectories. The denominator is the total probability of failure, which we denote as $\mu$. Due to the high dimensionality of trajectories and the complexity of autonomous systems, exact computation of this quantity is intractable. Instead, we rely on black-box evaluations of the system to estimate this probability.

\subsection{Failure Probability Estimation}

Failure probability estimation is the problem of estimating the denominator of \cref{eq:failure-distribution}, or the expected value:
\begin{equation}
    \mu = \mathbb{E}_{p(\tau)} \left[ \mathds{1} \{f(\tau) \geq \gamma \} \right]
\end{equation}
A simple approach to estimate $\mu$ is using Monte Carlo (MC) sampling, which uses $N$ independent samples from $p(\tau)$ to compute the empirical estimate
\begin{equation}
    \hat{\mu}_{\text{MC}} = \frac{1}{N} \sum_{i=1}^N \mathds{1} \{f(\tau_i) \geq \gamma \}
\end{equation}
For very rare failures (e.g. $\mu \approx 10^{-9}$), standard Monte Carlo may require billions of samples, making direct estimation infeasible for complex systems.


\subsection{Importance Sampling}
Importance sampling can reduce the variance of failure probability estimates by focusing samples on regions of interest. IS draws samples from a proposal distribution $q(\tau)$, and computes an estimate
\begin{equation}
    \label{eq:is-estimate}
    \hat{\mu}_{\text{IS}} = \frac{1}{N} \sum_{i=1}^N w(\tau_i) \mathds{1} \{f(\tau_i) \geq \gamma\}
\end{equation}
where the samples are weighted according to their importance weight $w(\tau) = p(\tau)/q(\tau)$. The minimum variance importance sampling distribution is the failure distribution itself, meaning that samples are only drawn from the failure region with likelihood proportional to $p(\tau)$. The estimator is unbiased if $q(\tau) > 0$ whenever $ \mathds{1} \{f(\tau_i) \geq \gamma\} p(\tau) > 0$.

The key challenge of importance sampling is designing a good proposal distribution. The variance of the estimator increases roughly exponentially with the problem dimension, but it can be reduced by a well-designed proposal distribution. However, modeling the complex $D_xT$-dimensional trajectory distributions following previous IS approaches is extremely difficult.

\section{Methods}

Next, we introduce our method for failure probability estimation in sequential systems. We propose a sequential state-based IS proposal, discuss proposal optimization, and detail our algorithm. 

\subsection{Sequential Proposal Distribution}

To address the challenges of fitting a complex, high-dimensional proposal distribution, we decompose the problem using a proposal applied at each timestep. Since the system behavior and disturbance distribution may depend on the system state, we condition the distribution on the current state. These proposals take the form $q_{\theta}(x \mid s)$ with parameters $\theta$. We can draw samples from $q_{\theta}(\tau)$ by sampling an initial state $s_1 \sim p(s_1)$ and simulating each timestep:
\begin{equation}
    \label{eq:q-traj-simulate}
    x_t \sim q_{\theta}(\cdot \mid s_t), \quad o_t 
    \sim p(\cdot \mid s_t, x_t), \quad a_t \sim \pi(\cdot \mid o_t, x_t), \quad s_{t+1} \sim p(\cdot \mid s_t, a_t, x_t)
\end{equation}
System trajectories sampled using the proposal $q_{\theta}(x \mid s)$ have the density:
\begin{equation}
    \label{eq:q-density}
    q_{\theta}(\tau) = p(s_1) \prod_{t=1}^T p(s_{t+1} \mid s_t, a_t, x_t) p(o_t \mid s_t, x_t) \pi(a_t \mid o_t, x_t) q_{\theta}(x_t \mid s_t)
\end{equation}
In computing the importance weight as the ratio of \cref{eq:nominal-density} to \cref{eq:q-density}, all terms cancel except the disturbance distribution and the proposal. The importance weight reduces to:
\begin{equation}
    w(\tau) = \prod_{t=1}^T \frac{d(x_t \mid s_t)}{q_{\theta}(x_t \mid s_t)}
\end{equation}
This formulation reduces the problem of learning a proposal over full disturbance trajectories to learning a proposal over a single timestep given the current state. However, obtaining a reliable IS estimate of the failure probability still requires careful design of this proposal.

\subsection{Proposal Optimization}
The optimal importance sampling distribution is known to be proportional to the true distribution over failure trajectories. Therefore, we propose to learn the proposal parameters by minimizing the forward KL divergence between the proposal and the failure distribution:
\begin{equation}
    \label{eq:full-kl-objective}
    \theta^* = \argmin_{\theta} D_{KL}\left(p \left(\tau \mid \mathds{1}\{f(\tau) > \gamma\} \right) \mid \mid q_{\theta}\left(\tau\right) \right)
\end{equation}
Minimizing the forward KL divergence is desirable for importance sampling because its mass covering properties help ensure the IS estimator remains unbiased \citep{rubinstein2004cross}. This objective is equivalent up to a constant of proportionality to minimizing the cross-entropy:
\begin{equation}
    \label{eq:full-ce-objective}
    \min_{\theta} \mathcal{L}(\theta)  = \min_{\theta} \mathbb{E}_{p(\tau \mid \mathds{1} \{ f(\tau) \geq \gamma\})} \left[-\log q_{\theta}(\tau) \right]
\end{equation}
However, this formulation creates a chicken-and-egg problem, since our goal is to approximate $p(\tau \mid \mathds{1} \{ f(\tau) \geq \gamma\})$ with $q_{\theta}(\tau)$, but the cross-entropy involves an expectation over $p(\tau \mid \mathds{1} \{ f(\tau) \geq \gamma\})$. A common approach optimizes an IS estimate of the expectation. Unfortunately, the variance of the IS estimate increases rapidly with problem dimension, making optimization very difficult especially when the proposal is far from the target.

Building on previous work in variational Bayesian inference, we optimize the objective in \cref{eq:full-ce-objective} using Markov score ascent methods \citep{naesseth2020msc, kim2022msa}. Markov score ascent methods can achieve much lower variance estimates of the objective. The key idea of Markov score ascent methods is to estimate the cross-entropy using samples from an MCMC kernel that is ergodic with respect to the target distribution. The kernel is chosen to directly use the approximation of the target distribution $q_{\theta}(\tau)$. With repeated application of the kernel, the samples will be approximately distributed according to the target. Given $N$ trajectories approximately distributed according to the target, we may compute a Monte Carlo estimate of the objective:
\begin{equation}
    \mathcal{L}(\theta) \approx \frac{1}{N} \sum_{n=1}^N - \log q_{\theta}(\tau_n)
\end{equation}
 Expanding the objective and removing terms that do not depend on $\theta$, the loss can be computed as
\begin{equation}
    \label{eq:expanded-loss}
    \mathcal{L}(\theta) \approx \frac{1}{N} \sum_{n=1}^N  \sum_{t=1}^T- \log q_{\theta}(x_{t, n} \mid s_{t,n})
\end{equation}
where subscript $t,n$ denotes timestep $t$ in trajectory $n$. This estimated cross-entropy objective is optimized using stochastic gradient descent.

In this work, we construct an MCMC kernel by performing a single step of Independent Metropolis-Hastings \citep[IMH,][]{hastings1970monte}. Given an initial trajectory $\tau$, we can generate a new sample from IMH by first sampling a proposal trajectory $\tau' \sim q_{\theta}(\cdot)$. IMH either accepts the new trajectory $\tau'$ according to the Metropolis-Hastings acceptance probability $a$:
\begin{equation}
    \label{eq:acceptance-ratio}
    a(\tau, \tau') = \min \left(1, \frac{p(\tau' \mid \mathds{1} \{ f(\tau') \geq \gamma\}) q_{\theta}(\tau)}{p(\tau \mid \mathds{1} \{ f(\tau) \geq \gamma\}) q_{\theta}(\tau')}\right) = \min \left(1, \frac{w(\tau')}{w(\tau)}\right)
\end{equation}
Otherwise, IMH rejects the proposal and $\tau$ remains the same.

One challenge in applying IMH to sample from the true failure distribution is that the failure distribution applies zero density to safe trajectories. Since failures are rare, many sampled trajectories may be safe. In turn, the acceptance probabilities will be zero, making it very difficult for the MCMC step to move trajectories closer to the failure distribution.


\subsection{Approximating the Failure Distribution}
\begin{wrapfigure}{R}{0.4\textwidth}
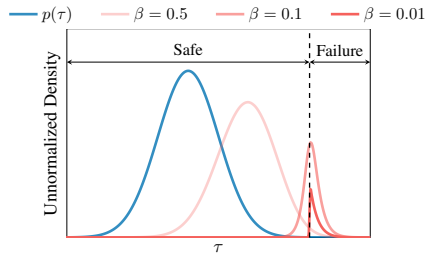

  \begin{center}
    \includestandalone[width=\linewidth]{smooth_target}
  \end{center}
  \caption{Illustration of $\tilde{p}_{\beta}(\tau)$.}
  \label{fig:smooth-target}
\end{wrapfigure}
To address the issue of low acceptance rates, we use a smooth approximation of the discontinuous failure distribution. Specifically, we approximate the indicator function in the target by $P_{\beta}(f(\tau) - \gamma)$ where $P_\beta$ is the cumulative distribution function of a logistic distribution with zero mean and scale $\beta$. This approximation replaces the discontinuous indicator with a smooth logistic curve from zero to one. The approximate failure distribution $\tilde{p}(\tau)$ has the unnormalized density
\begin{equation}
    \label{eq:approx-density}
    \tilde{p}(\tau) \propto p(\tau)P_{\beta}(f(\tau) - \gamma)
\end{equation}
which relaxes the failure distribution, enabling stable sampling during optimization. This approximation is visualized in \cref{fig:smooth-target} for a few values of $\beta$.


Using the approximate posterior, our objective becomes:
\begin{equation}
    \label{eq:approx-ce-objective}
    \mathcal{L}(\theta) = \mathbb{E}_{\tilde{p}(\tau)} \left[-\log q_{\theta}(\tau) \right]
\end{equation}
We can now apply IMH to generate proposal samples for Markov score ascent by computing the acceptance probability using the unnormalized importance weights under the approximate failure distribution $\tilde{w}(\tau) = \tilde{p}(\tau)/q_{\theta}(\tau)$. 
Note that we can still compute the acceptance probability using $\tilde{w}(\tau)$ without knowing the normalizing constant of the approximate failure distribution, because the constant values in the numerator and denominator in the ratio of \cref{eq:acceptance-ratio} cancel.

\subsection{Algorithm Details}
The complete algorithm, which we call State-dependent Proposal Adaptive Importance Sampling (SPAIS), is shown in \cref{alg:method}. The algorithm maintains a set of $N$ trajectories that will be updated using the IMH kernel and a buffer to store evaluations $f(\tau)$ and the corresponding IS weight of samples from $q_\theta(\tau)$. At each iteration, the algorithm samples N new proposal trajectories $\tau'$. These samples are first added to the IS buffer. Next, they are used in the step to update the trajectory set. Each proposed trajectory is accepted or rejected according using the MH acceptance probability. At the end of each iteration, we update the proposal by fitting the proposal to the updated set of trajectories. After a fixed number of iterations, the algorithm returns an IS estimate of the failure probability using \cref{eq:is-estimate}.

In practice, we parameterize $q_{\theta}(x \mid s)$ as a multivariate Gaussian
\begin{equation*}
    q_{\theta}(x \mid s) = \mathcal{N} (x \mid \mu_{\theta}(s), \Sigma_{\theta}(s))
\end{equation*}
where $\mu_{\theta}(\cdot)$ and $\Sigma_{\theta}(\cdot)$ are both neural networks with two hidden layers of size $64$ and $32$ respectively.  We find using any small value of $\beta$ in the range $[10^{-4}, 10^{-2}]$ gives good performance. We use $\beta=10^{-2}$ for all evaluations. For each problem, we normalize the state space features and pretrain $q_{\theta}(x \mid s)$ to match $d(x \mid s)$.

\input{algorithm}

\section{Experiments}
In this section, we describe our experiments to evaluate the proposed approach. We describe baseline estimation methods,  evaluation metrics, and introduce four example safety validation problems.

\subsection{Evaluation Metrics and Baselines}
We evaluate the bias and variance of the estimated probability of failure using the relative error $\epsilon_{\text{rel}}$ and absolute relative error $\epsilon_{\text{abs}}$, respectively:
\begin{equation*}
    \epsilon_{\text{rel}} = (\hat{\mu} - \mu)/\mu, \quad \epsilon_{\text{abs}} = \vert\hat{\mu} - \mu\vert/\mu
\end{equation*}
We obtain an estimate of the ground truth failure probability $\mu$ using $10^7$ Monte Carlo samples. Each metric is averaged over $10$ trials with $50000$ samples per trial to arrive at estimates of the empirical bias and variance for each method.

We compare the method against three baseline methods. The simplest baseline is Monte Carlo estimation. We also compare against a variant of the cross-entropy method (CEM) that optimizes a state-independent proposal distribution applied at each timestep. Finally, we evaluate against PG-AIS, which learns a reinforcement learning policy for importance sampling \citep{corso2022deep}.

\subsection{Validation Problems}
\begin{wraptable}{r}{0.4\textwidth}
    \centering
    \vspace{-0mm}
    \resizebox{0.4\textwidth}{!}{
    \begin{tabular}{@{}l r r r r @{}}
    \toprule
    \text{Problem} & $\mu$ & $D_s$ & $D_x$ & $T$ \\
    \midrule
    Pendulum & \num{1.96e-5} & $2$ & $1$ & $20$ \\
    Crosswalk & \num{5.90e-5} & $8$ & $5$ & $100$ \\
    Collision Avoidance & \num{2.23e-5} & $12$ & $1$ & $40$ \\
    F-16 GCAS & \num{6.59e-5} & $13$ & $6$ & $200$ \\
    \bottomrule
    \end{tabular}
    }
    \caption{Failure probability $\mu$, state size $D_s$, disturbance size $D_x$, and horizon $T$ for each problem.}
    \label{tab:problem-summary}
\end{wraptable} 

\begin{figure}[!t]
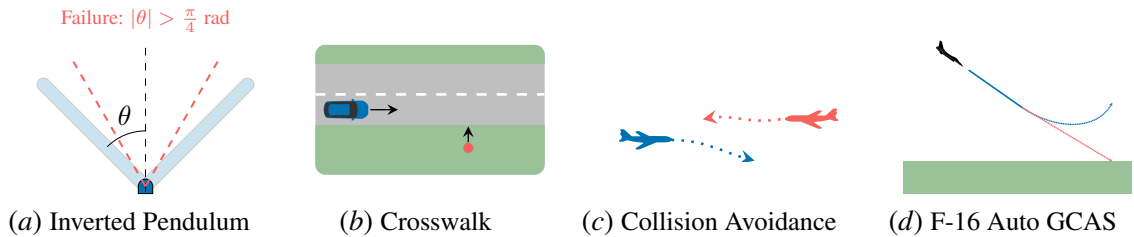

    \subfigure[\small Inverted Pendulum]{
        \label{fig:pendulum}
        \includestandalone[width=0.2\linewidth]{pendulum}
    }
    \hfill
    \subfigure[\small Crosswalk]{
    \label{fig:systems-jaywalk}
    \raisebox{0.15\height} {\includestandalone[width=0.2\linewidth]{crosswalk}}
    }
    \hfill
    \subfigure[\small Collision Avoidance]{
    \label{fig:cas}
    \raisebox{0.6\height}{
    \includestandalone[width=0.2\linewidth]{acas}
    }
    }
    \hfill
    \subfigure[\small F-16 Auto GCAS]{
    \label{fig:f16}
    \includestandalone[width=0.2\linewidth]{f16}
    }
    \caption{The environments used in failure probability estimation experiments.}
    \label{fig:systems}
\end{figure}

We demonstrate SPAIS on four systems illustrated in \cref{fig:systems}. Each problem's properties are reported in \cref{tab:problem-summary}.

\paragraph{Inverted Pendulum} We evaluate an underactuated inverted pendulum system with a nonlinear rule-based control policy. The system is subjected to additive torque disturbances over 20 timesteps. We define failure to occur when the magnitude of the pendulum's angle from the vertical exceeds $\SI[parse-numbers=false]{\pi/4}{rad}$. This problem has two distinct failure modes corresponding to leftward and rightward falls.

\paragraph{Crosswalk} In the second experiment we evaluate an autonomous vehicle (AV) approaching a pedestrian at a crosswalk inspired by \citet{koren2018adaptive}. The AV uses the Intelligent Driver Model \citep{treiber2000IDM} for longitudinal control, maintaining safe distances from obstacles.  We adversarially control the pedestrian's 2D acceleration, perceived position and speed as detected by the AV. Failures occur when the AV collides with the pedestrian.

\paragraph{Aircraft Collision Avoidance} The third experiment evaluates an aircraft collision avoidance system based on the AVOIDDS dataset and simulator \citep{smyers2023avoidds}. The system detects nearby intruding aircraft and recommends pilot maneuvers (e.g. climb or descend) to avoid midair collisions with the ownship. We evaluate the collision avoidance policy in scenarios with perfect perception of a single intruding aircraft. The intruder's initial state is randomly sampled according to the AVOIDDS encounter model. We adversarially control the intruder's vertical rate to cause midair collisions with the ownship.

\paragraph{F-$\mathbf{16}$ Ground Collision Avoidance} Finally, we consider a model of the ground collision avoidance system (GCAS) for the F-16 aircraft adapted from \citet{heidlauf2018verification}. The GCAS detects impending ground collisions and executes a series of maneuvers to prevent collision. We add disturbances to observations of orientation and angular velocity over $200$ timesteps. The aircraft starts at an altitude of $\SI{578}{ft}$, pitched down $\SI{27}{deg}$, and is rolled $\SI{22}{deg}$. The controller first rolls the wings level and then pitches up to avoid collision.

\section{Results}

The relative and absolute error for each method across all problems are reported in \cref{tab:results}. The proposed SPAIS algorithm achieves lower relative and absolute error on all four validation problems. The MC estimator generally has low bias, but high variance. The PG-AIS baseline method tends to have large variance, indicating that it struggles to represent the failure distribution. CEM exhibits a bias, which may be due to mode collapse where the proposal does not cover all failure modes or due to the proposal being spread very wide leading to large importance weights.

\begin{table}[!t]
    \centering
    \begin{adjustbox}{max width=\textwidth}
    \begin{tabular}{
        @{} 
        l 
        S[table-format=1.2(3)]
        S[table-format=1.2(3)]
        S[table-format=1.2(3)]
        S[table-format=1.2(3)]
        S[table-format=1.2(3)]
        S[table-format=1.2(3)]
        S[table-format=1.2(3)]
        S[table-format=1.2(3)]
        @{}
    }
    \toprule
          & \multicolumn{2}{c}{$\text{Pendulum}$} 
          & \multicolumn{2}{c}{$\text{Crosswalk}$} 
          & \multicolumn{2}{c}{$\text{Collision Avoidance}$} 
          & \multicolumn{2}{c}{$\text{F-16 GCAS}$} \\
         Method & $\epsilon_{\text{rel}}$ & $\epsilon_{\text{abs}}$ 
                & $\epsilon_{\text{rel}}$ & $\epsilon_{\text{abs}}$ 
                & $\epsilon_{\text{rel}}$ & $\epsilon_{\text{abs}}$ 
                & $\epsilon_{\text{rel}}$ & $\epsilon_{\text{abs}}$ \\
    \midrule
    MC      & -0.39(99)   & 0.81(65)    & 0.15(36)    & 0.28(27)    & 0.30(55)    & 0.54(26)    & 0.39(88)    & 0.74(70)    \\
    CEM     & -0.54(5)    & 0.53(5)     & -0.22(11)   & 0.22(11)    & -0.18(50)   & 0.26(43)    & 0.63(175)   & 1.13(109)   \\
    PG-AIS  & -0.41(28)    & 0.47(16)     & -0.26(55)   & 0.41(44)    & -0.91(10)   & 0.91(10)    & -0.91(45)   & 0.94(58)    \\
    SPAIS   & \bfseries -0.04(6)    & \bfseries 0.06(3) & \bfseries -0.02(13)   & \bfseries 0.09(9)     & \bfseries -0.05(27)   & \bfseries 0.18(21)    & \bfseries 0.06(10)     & \bfseries 0.09(7)     \\
    \bottomrule
    \end{tabular}
    \end{adjustbox}
    \caption{\small Relative and absolute error for each method across all four validation problems.}
    \label{tab:results}
\end{table}

\begin{figure}[!t]
    \subfigure[\small Inverted Pendulum]{
        \label{fig:pendulum-states}
        \includegraphics[width=0.22\linewidth]{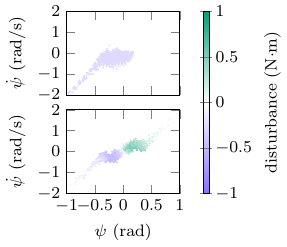}
    }
    \hfill
    \subfigure[\small Crosswalk]{
        \label{fig:crosswalk-states}
        \includegraphics[width=0.22\linewidth]{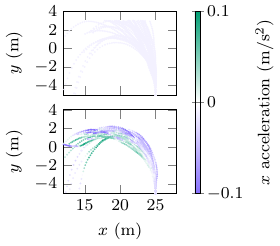}
    }
    \hfill
    \subfigure[\small Collision Avoidance]{
        \label{fig:cas-states}
        \includegraphics[width=0.22\linewidth]{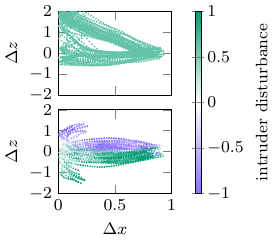}
    }
    \hfill
    \subfigure[\small F-16 GCAS]{
        \label{fig:f16-states}
        \includegraphics[width=0.22\linewidth]{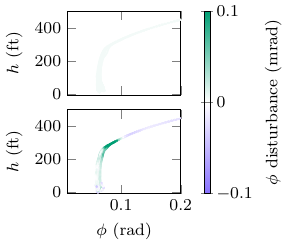}
    }
    \caption{Visited states and applied disturbances by CEM (above) and SPAIS (below) after training.}
    \label{fig:state-plots}
\end{figure}

We visualize sampled system trajectories from the CEM and SPAIS proposals for each problem in \cref{fig:state-plots}. Each plot shows a system trajectories in state space with points colored according to a mean input disturbance sampled from the proposal.


\paragraph{Inverted Pendulum} In \cref{fig:pendulum-states}, we plot the pendulum angle $\psi$ against angular rate $\dot{\psi}$, and color the points according to the torque disturbance. SPAIS generally adds positive torque disturbances for positive angles, and vice versa. In contrast, the CEM proposal only samples negative disturbances and thus only samples from one of the two failure modes.


\paragraph{Crosswalk} CEM and SPAIS proposals for the crosswalk problem are shown in \cref{fig:crosswalk-states}. This figure shows the $2$D position of the pedestrian with points colored according to the controlled pedestrian $x$-axis acceleration. Both proposals steer the pedestrian into the street and towards the oncoming AV. The SPAIS proposal is more structured around an important failure mode. The trajectories are more consistent, and use some deceleration to keep the pedestrian in front of the AV for longer. In contrast, the CEM proposal is more spread out, potentially contributing to its larger bias.


\paragraph{Aircraft Collision Avoidance} Figure \ref{fig:cas-states} shows the proposals for the aircraft collision avoidance problem, plotting normalized vertical separation $\Delta z$ against horizontal distance $\Delta x$ of the ownship and intruder. The CEM proposal exhibits mode collapse, favoring intruder trajectories below the ownship. The SPAIS proposal symmetrically samples trajectories above and below the ownship, contributing to reduced estimation bias.


\paragraph{F-16 GCAS} Figure \ref{fig:f16-states} shows F-16 GCAS trajectories of altitude $h$ against roll angle $\phi$ with points colored by the observed roll angle disturbance. SPAIS increases the error in the observed roll angle just before the controller begins to pull up, causing a delay in the maneuver and more ground collisions. The mean CEM disturbance is small, suggesting that the proposal must be wide to capture the failure trajectories discovered by SPAIS.

\section{Conclusion}
Accurately estimating failure probabilities is a critical step in developing autonomous systems in safety-critical domains.  We propose a method for failure probability estimation that takes advantage of the sequential nature of autonomous systems to construct flexible IS proposals that are optimized using Markov score ascent. Using a relatively small number of samples, our results show that SPAIS achieves under $10\%$ empirical bias with at least $2$ times lower empirical variance than baselines. Future work will investigate using more flexible state-based proposals to further improve estimation.

\acks{This material is based upon work supported by the National Science Foundation Graduate Research Fellowship under Grant No. DGE-2146755. Any opinions, findings, conclusions or recommendations expressed in this material are those of the authors and do not necessarily reflect the views of the National Science Foundation.}

\bibliography{references}

\end{document}

%% file: preamble/preamble.tex
\usepackage{booktabs}
\usepackage[detect-mode=true]{siunitx}
\usepackage{wrapfig}
\usepackage{xcolor}
\usepackage{graphicx}
\graphicspath{{./figures/}}
\usepackage{amsmath}
\usepackage{amsfonts}
\usepackage{cleveref}
\usepackage{vector} 
\usepackage[font=small]{caption}
\usepackage{adjustbox}
\usepackage{longtable, booktabs}
\usepackage[flushleft]{threeparttable}
\usepackage{multirow}
\usepackage{lipsum}
\usepackage{dsfont} 
\usepackage{colortbl}
\usepackage[mode=buildnew]{standalone}

\definecolor{juliafunccolor}{HTML}{2e51a2}
\definecolor{juliaparenscolor}{HTML}{f2d71c}
\definecolor{pastelRed}{RGB}{245, 97, 92}
\definecolor{pastelBlue}{RGB}{0, 114, 178}
\definecolor{pastelGreen}{RGB}{0, 158, 115}
\definecolor{pastelPurple}{RGB}{135, 112, 254}

\definecolor{newcolor}{HTML}{000000}

\usepackage{algorithmicx}
\usepackage{algorithm}
\usepackage{algpseudocode}

\usepackage{tikz}
\usepackage{pgfplots}
\pgfplotsset{compat=newest}

\usetikzlibrary{calc}
\usetikzlibrary{shapes.geometric}
\usetikzlibrary{external}
\usetikzlibrary{patterns}
\usetikzlibrary{shapes,arrows,fit}
\usetikzlibrary{positioning}
\usetikzlibrary{arrows.meta, calc, shapes}
\usetikzlibrary{graphs}
\usetikzlibrary{decorations.pathmorphing}
\usetikzlibrary{decorations.pathreplacing}
\usetikzlibrary{backgrounds}
\usetikzlibrary{shadows}
\usepgfplotslibrary{fillbetween}

\usepackage{pagecolor}

\definecolor{blues1}{RGB}{198, 219, 239}
\definecolor{blues2}{RGB}{158, 202, 225}
\definecolor{blues3}{RGB}{107, 174, 214}
\definecolor{blues4}{RGB}{49, 130, 189}
\definecolor{blues5}{RGB}{8, 81, 156}

\definecolor{grays1}{RGB}{219, 219, 219}
\definecolor{grays2}{RGB}{202, 202, 202}
\definecolor{grays3}{RGB}{174, 174, 174}
\definecolor{grays4}{RGB}{130, 130, 130}
\definecolor{grays5}{RGB}{81, 81, 81}

\usetikzlibrary{calc,tikzmark,positioning,shadows,arrows}

\makeatletter

\definecolor{pastelred}{RGB}{245,97,92}
\definecolor{steelblue}{rgb}{0.27,0.51,0.71}

\def\pgf@body@color{black!30}
\def\pgf@window@color{black!10}

\pgfdeclareshape{sedan top}{
\inheritsavedanchors[from=coordinate]
\inheritanchor[from=coordinate]{center}
\backgroundpath{
\begin{pgfscope}
    \pgftransformshift{\centerpoint}
    \pgfmathdivide{\pgfkeysvalueof{/pgf/minimum width}}{1pt}
    \pgftransformscale{\pgfmathresult}

    \pgfpathmoveto{\pgfpoint{-0.3764698682720557}{0.16169217949171047}}
    \pgfpathlineto{\pgfpoint{0.3371231697857909}{0.17694315802976227}}
    \pgfpathlineto{\pgfpoint{0.36786879148692225}{0.16589767639860994}}
    \pgfpathlineto{\pgfpoint{0.4064052148177093}{0.13879466238659263}}
    \pgfpathlineto{\pgfpoint{0.4393446579325617}{0.10613070348195595}}
    \pgfpathlineto{\pgfpoint{0.45329950694358345}{0.07840237877506098}}
    \pgfpathlineto{\pgfpoint{0.45329950694358345}{-0.07840334453180331}}
    \pgfpathlineto{\pgfpoint{0.43933991312769727}{-0.1061282429017343}}
    \pgfpathlineto{\pgfpoint{0.4063926179906356}{-0.1387866550035162}}
    \pgfpathlineto{\pgfpoint{0.3678546410511762}{-0.16588533150807783}}
    \pgfpathlineto{\pgfpoint{0.3371231697857909}{-0.17693102308634803}}
    \pgfpathlineto{\pgfpoint{-0.3764698682720557}{-0.16169356514268857}}
    \pgfpathlineto{\pgfpoint{-0.3944099628675368}{-0.15772328458879273}}
    \pgfpathlineto{\pgfpoint{-0.4099758704803143}{-0.14724966111630228}}
    \pgfpathlineto{\pgfpoint{-0.43396985110493147}{-0.11051053207766995}}
    \pgfpathlineto{\pgfpoint{-0.4484212506633208}{-0.05891246295317149}}
    \pgfpathlineto{\pgfpoint{-0.4532995069435834}{0.00010826972869789077}}
    \pgfpathlineto{\pgfpoint{-0.4485740585736087}{0.059115381041558314}}
    \pgfpathlineto{\pgfpoint{-0.4342143437613922}{0.11067259025797227}}
    \pgfpathlineto{\pgfpoint{-0.41018980239450586}{0.14734361665050213}}
    \pgfpathlineto{\pgfpoint{-0.39454366978903876}{0.15777297487265582}}
    \pgfpathlineto{\pgfpoint{-0.3764698682720557}{0.16169217949171047}}
    \pgfpathclose

    \pgfpathmoveto{\pgfpoint{0.13271865580424752}{0.21099658055215992}}
    \pgfpathlineto{\pgfpoint{0.14640536035629512}{0.21031786350943238}}
    \pgfpathlineto{\pgfpoint{0.1537574564883996}{0.19881177469739017}}
    \pgfpathlineto{\pgfpoint{0.16526045907780879}{0.16238122593263837}}
    \pgfpathlineto{\pgfpoint{0.14705384501404595}{0.15750667374937677}}
    \pgfpathlineto{\pgfpoint{0.13271865580424752}{0.21099658055215992}}
    \pgfpathclose

    \pgfpathmoveto{\pgfpoint{0.13271865580424752}{-0.21099658055215995}}
    \pgfpathlineto{\pgfpoint{0.14705384501404595}{-0.15751099866000545}}
    \pgfpathlineto{\pgfpoint{0.16526045907780879}{-0.16238219168938073}}
    \pgfpathlineto{\pgfpoint{0.1537574564883996}{-0.19880160066005706}}
    \pgfpathlineto{\pgfpoint{0.14640536035629512}{-0.21030964827870924}}
    \pgfpathlineto{\pgfpoint{0.13271865580424752}{-0.21099658055215995}}
    \pgfpathclose
    \color{\pgf@body@color}
    \pgfusepath{fill,draw}
\end{pgfscope}
}
    
\beforebackgroundpath{
\begin{pgfscope}
    \pgftransformshift{\centerpoint}
    \pgfmathdivide{\pgfkeysvalueof{/pgf/minimum width}}{1pt}
    \pgftransformscale{\pgfmathresult}
    
    \pgfpathmoveto{\pgfpoint{-0.3278343587292163}{-0.14804700247958225}}
    \pgfpathlineto{\pgfpoint{-0.2234184193273709}{-0.15031023241047897}}
    \pgfpathlineto{\pgfpoint{-0.21525116152061965}{-0.14108410631462961}}
    \pgfpathlineto{\pgfpoint{-0.21003597013857514}{-0.13038163208575007}}
    \pgfpathlineto{\pgfpoint{-0.28007422369443574}{-0.13072059170758982}}
    \pgfpathlineto{\pgfpoint{-0.30920043509746165}{-0.1357313491830563}}
    \pgfpathlineto{\pgfpoint{-0.3278343587292163}{-0.14804700247958225}}
    \pgfpathclose
    
    \pgfpathmoveto{\pgfpoint{-0.20348981900264201}{-0.15067134145325653}}
    \pgfpathlineto{\pgfpoint{-0.06450146780296512}{-0.1538205482216657}}
    \pgfpathlineto{\pgfpoint{-0.06439150590049697}{-0.1433681755964485}}
    \pgfpathlineto{\pgfpoint{-0.06919084662824423}{-0.13038163208575007}}
    \pgfpathlineto{\pgfpoint{-0.19919598054747503}{-0.13038163208575007}}
    \pgfpathlineto{\pgfpoint{-0.20370480485136538}{-0.1408899052305777}}
    \pgfpathlineto{\pgfpoint{-0.20348981900264201}{-0.15067134145325653}}
    \pgfpathclose
    
    \pgfpathmoveto{\pgfpoint{-0.045098155167206846}{-0.1540846616959763}}
    \pgfpathlineto{\pgfpoint{0.15615715204592737}{-0.1585443583742794}}
    \pgfpathlineto{\pgfpoint{0.1406284134709018}{-0.14934034810782904}}
    \pgfpathlineto{\pgfpoint{0.11302876535256408}{-0.1409152584445346}}
    \pgfpathlineto{\pgfpoint{0.025565845773536638}{-0.13038163208575007}}
    \pgfpathlineto{\pgfpoint{-0.057380061564002535}{-0.13038163208575007}}
    \pgfpathlineto{\pgfpoint{-0.05322499815376351}{-0.14451459183370835}}
    \pgfpathlineto{\pgfpoint{-0.045098155167206846}{-0.1540846616959763}}
    \pgfpathclose
    
    \pgfpathmoveto{\pgfpoint{-0.40326793829427443}{0.13060698932209738}}
    \pgfpathlineto{\pgfpoint{-0.3824075926603322}{0.1357263398448233}}
    \pgfpathlineto{\pgfpoint{-0.3588179344937687}{0.12593398637201397}}
    \pgfpathlineto{\pgfpoint{-0.3514240168957798}{0.12001473733011214}}
    \pgfpathlineto{\pgfpoint{-0.34496142471276847}{0.11392921017083832}}
    \pgfpathlineto{\pgfpoint{-0.3529478130774541}{0.11007500098054127}}
    \pgfpathlineto{\pgfpoint{-0.3657244094702585}{0.09329891199185718}}
    \pgfpathlineto{\pgfpoint{-0.37254805610979685}{0.0726124571577818}}
    \pgfpathlineto{\pgfpoint{-0.3805180685992867}{0.000180701484371307}}
    \pgfpathlineto{\pgfpoint{-0.37254805610979685}{-0.0722854855163738}}
    \pgfpathlineto{\pgfpoint{-0.3657244094702585}{-0.09297209991025879}}
    \pgfpathlineto{\pgfpoint{-0.3529478130774541}{-0.10974802933913329}}
    \pgfpathlineto{\pgfpoint{-0.34496142471276847}{-0.11360265842366608}}
    \pgfpathlineto{\pgfpoint{-0.3514240168957798}{-0.11968776568870415}}
    \pgfpathlineto{\pgfpoint{-0.3588179344937687}{-0.125607014730606}}
    \pgfpathlineto{\pgfpoint{-0.3824075926603322}{-0.1353993682034153}}
    \pgfpathlineto{\pgfpoint{-0.40326793829427443}{-0.13026868053632312}}
    \pgfpathlineto{\pgfpoint{-0.4222685883138931}{-0.10640394125801027}}
    \pgfpathlineto{\pgfpoint{-0.43333610263538563}{-0.07768028665777657}}
    \pgfpathlineto{\pgfpoint{-0.44306093711508027}{1.652283817825885e-5}}
    \pgfpathlineto{\pgfpoint{-0.44306093711508027}{0.00034488013056435516}}
    \pgfpathlineto{\pgfpoint{-0.43333610263538563}{0.07800725829918456}}
    \pgfpathlineto{\pgfpoint{-0.4222685883138931}{0.10673106826028549}}
    \pgfpathlineto{\pgfpoint{-0.40326793829427443}{0.13060698932209738}}
    \pgfpathclose
    
    \pgfpathmoveto{\pgfpoint{0.08894048278100244}{0.12330964739833968}}
    \pgfpathlineto{\pgfpoint{0.19794922533389572}{0.14278056300605982}}
    \pgfpathlineto{\pgfpoint{0.20644788466624256}{0.13751550918351546}}
    \pgfpathlineto{\pgfpoint{0.222695482172937}{0.06873333144353315}}
    \pgfpathlineto{\pgfpoint{0.22845193821962387}{-5.4754208346744125e-5}}
    \pgfpathlineto{\pgfpoint{0.2232262704763964}{-0.06884897871395386}}
    \pgfpathlineto{\pgfpoint{0.2065276645710423}{-0.1376495814130026}}
    \pgfpathlineto{\pgfpoint{0.19806679571991634}{-0.14278194865703794}}
    \pgfpathlineto{\pgfpoint{0.08894048278100244}{-0.12328122055857689}}
    \pgfpathlineto{\pgfpoint{0.08393954243276867}{-0.1201782021561044}}
    \pgfpathlineto{\pgfpoint{0.08264206924418414}{-0.11443908774135557}}
    \pgfpathlineto{\pgfpoint{0.09009859307272902}{-0.05721800915724597}}
    \pgfpathlineto{\pgfpoint{0.09258411501205184}{9.964090215246757e-6}}
    \pgfpathlineto{\pgfpoint{0.09009859307272902}{0.05724005780356717}}
    \pgfpathlineto{\pgfpoint{0.08264206924418414}{0.11446751458111835}}
    \pgfpathlineto{\pgfpoint{0.08393954243276867}{0.12019487195726514}}
    \pgfpathlineto{\pgfpoint{0.08894048278100244}{0.12330964739833968}}
    \pgfpathclose
    
    \pgfpathmoveto{\pgfpoint{-0.045098155167206846}{0.15407991689111186}}
    \pgfpathlineto{\pgfpoint{0.15615715204592737}{0.1585429727233013}}
    \pgfpathlineto{\pgfpoint{0.1406284134709018}{0.14934404737604628}}
    \pgfpathlineto{\pgfpoint{0.11302876535256408}{0.14091770222898686}}
    \pgfpathlineto{\pgfpoint{0.025565845773536638}{0.13038066632900772}}
    \pgfpathlineto{\pgfpoint{-0.057380061564002535}{0.13038066632900772}}
    \pgfpathlineto{\pgfpoint{-0.05322499815376351}{0.14451850524798585}}
    \pgfpathlineto{\pgfpoint{-0.045098155167206846}{0.15407991689111186}}
    \pgfpathclose
    
    \pgfpathmoveto{\pgfpoint{-0.20348981900264201}{0.15066995580227843}}
    \pgfpathlineto{\pgfpoint{-0.06450146780296512}{0.1538191625706876}}
    \pgfpathlineto{\pgfpoint{-0.06439150590049697}{0.14336038235943221}}
    \pgfpathlineto{\pgfpoint{-0.06919084662824423}{0.13038066632900772}}
    \pgfpathlineto{\pgfpoint{-0.19919598054747503}{0.13038066632900772}}
    \pgfpathlineto{\pgfpoint{-0.20370480485136538}{0.14089376825754693}}
    \pgfpathlineto{\pgfpoint{-0.20348981900264201}{0.15066995580227843}}
    \pgfpathclose
    
    \pgfpathmoveto{\pgfpoint{-0.3278343587292163}{0.14804561682860415}}
    \pgfpathlineto{\pgfpoint{-0.2234184193273709}{0.15030926665373662}}
    \pgfpathlineto{\pgfpoint{-0.21525116152061965}{0.14108251071653358}}
    \pgfpathlineto{\pgfpoint{-0.21003597013857514}{0.13038066632900772}}
    \pgfpathlineto{\pgfpoint{-0.28007422369443574}{0.13072445473455904}}
    \pgfpathlineto{\pgfpoint{-0.30920043509746165}{0.13573569508839667}}
    \pgfpathlineto{\pgfpoint{-0.3278343587292163}{0.14804561682860415}}
    \pgfpathclose
    \color{\pgf@window@color}
    \pgfusepath{fill,draw}
 \end{pgfscope}
 }
 }
\def\pgfsettopcolor#1{\def\pgf@body@color{#1}}%
\def\pgfsetbottomcolor#1{\def\pgf@window@color{#1}}%

\tikzoption{body color}{\pgfsettopcolor{#1}}
\tikzoption{window color}{\pgfsetbottomcolor{#1}}

\makeatother
\tikzset{%
  add/.style args={#1 and #2}{to path={%
 ($(\tikztostart)!-#1!(\tikztotarget)$)--($(\tikztotarget)!-#2!(\tikztostart)$)%
  \tikztonodes}}
} 

\tikzset{
    position/.style args={#1:#2 from #3}{
        at=(#3.#1), anchor=#1+180, shift=(#1:#2)
    }
}

\usepackage{aircraftshapes}

\DeclareMathOperator*{\argmin}{\arg\!\min}

\usetikzlibrary{pgfplots.colorbrewer}
\usepgfplotslibrary{groupplots}
\usepgfplotslibrary{polar}
\usepgfplotslibrary{smithchart}
\usepgfplotslibrary{statistics}
\usepgfplotslibrary{dateplot}
\usepgfplotslibrary{ternary}

\pgfplotsset{
    colormap={pastelPG}{
        color(0cm)=(pastelPurple)   
        color(0.5cm)=(white) 
        color(1cm)=(pastelGreen)    
    }
}

%% file: figures/showcase/showcase_standalone.tex
\begin{tikzpicture}




\node[] at (-13, 0) {
    \includestandalone{showcase/estimates}
};

\node[] at (-5.5, 0) {
    \includestandalone{showcase/group_trajectories}
};

\node[] at (0, 0) {
    \includestandalone{showcase/group_states}
};

\node[align=center] at (-2.75 , 0.25) {\large \textbf{Ours}};

\node[align=center] at (-2.75 , 4) {\large \textbf{Monte Carlo}};

\end{tikzpicture}

%% file: figures/smooth_target.tex
\newcommand{\normalPDF}[1]{1 / sqrt(2 * pi) * exp(-(#1)^2 / 2)}
\newcommand{\logisticCDF}[2]{1 / (1 + exp(-(#1) / #2))}
\definecolor{pastelRed}{RGB}{245, 97, 92}
\definecolor{pastelBlue}{RGB}{0, 114, 178}
\definecolor{pastelGreen}{RGB}{0, 158, 115}
\definecolor{pastelPurple}{RGB}{135, 112, 254}

\begin{tikzpicture}
    \pgfmathsetmacro{\scale}{0.01}  
    \pgfmathsetmacro{\mu}{4}     

    \begin{axis}[
        xlabel={$\tau$},
        ylabel={Unnormalized Density},
        domain=-6:6,
        samples=250,
        ymin=0, ymax=0.5,
        xmin=-4, xmax=6,
        ticks=none,
        legend style={draw=none, /tikz/every even column/.append style={column sep=0.5cm}},
        legend columns=-1,
        legend style={at={(axis description cs:0.5,1.0)}, anchor=south},
        width=8cm, height=6cm,
        enlargelimits=false,
    ]

        \addplot[pastelBlue, ultra thick, smooth, opacity=0.8] 
            {\normalPDF{x}};
        \addlegendentry{\small $p(\tau)$}

        \addplot[pastelRed, ultra thick, smooth, opacity=0.3] 
            {(\normalPDF{x} * \logisticCDF{x - \mu}{0.5})/0.003};
        \addlegendentry{\small $\beta=0.5$}

        \addplot[pastelRed, ultra thick, smooth, opacity=0.6]
            {(\normalPDF{x} * \logisticCDF{x - \mu}{0.1})/0.0003};
        \addlegendentry{\small $\beta=0.1$}
            
        \addplot[pastelRed, ultra thick, smooth] 
            {(\normalPDF{x} * \logisticCDF{x - \mu}{\scale})/0.001};
        \addlegendentry{\small $\beta=0.01$}

        \addplot[black, dashed, thick] coordinates {(4,0) (4,0.5)};

        \draw[stealth-stealth] (axis cs:-4,0.42) -- (axis cs:4,0.42) node[midway, above] {\small Safe};

        \draw[stealth-stealth] (axis cs:4,0.42) -- (axis cs:6,0.42) node[midway, above] {\small Failure};
        
    \end{axis}
\end{tikzpicture}

%% file: algorithm.tex
\begin{algorithm}[!t]
    \scriptsize
    \caption{\small State-dependent Proposal Adaptive Importance Sampling}
    \label{alg:method}
    \KwIn{nominal trajectory distribution \( p(\tau) \), evaluation function \( f(\tau) \), proposal \( q_{\theta_1}(x \mid s) \)}
    \KwIn{failure threshold \( \gamma \), number of particles \( N \), number of iterations \( N_{\text{iter}} \)}
    \KwOut{failure probability estimate $\hat{\mu}$}
    
    \SetKwFunction{Solve}{SPAIS}
    \SetKwFunction{MHAccept}{\textsc{MHAccept}}
    \SetKwFunction{Estimate}{ImportanceSamplingEstimate}
    \SetKwProg{Fn}{Function}{}{end}
    \SetAlgoLined
    \SetFuncSty{textproc}

    \Fn{\Solve{$p(\tau), f(\tau), q_{\theta_1}(x \mid s), \gamma, N, N_{\text{iter}}$}}{

        \( \tau_n \sim q_{\theta_1}(\tau) \) for \( n \in 1:N \)\; \Comment{Sample initial trajectories using \cref{eq:q-traj-simulate}}

        \( \mathcal{D} \leftarrow \{f(\tau_n), p(\tau_n) / q_{\theta_1}(\tau_n)\}_{n=1}^N \)\; \Comment{Initialize IS buffer}

        \( \tilde{w}_n \leftarrow p(\tau_n) P_{\beta}(f(\tau_n) - \gamma) / q_{\theta_1}(\tau_n) \) for \( n \in 1:N \)\; \Comment{Compute MH weights}

        \For{$k \in 1:N_{\text{iter}}$}{

            \( \tau'_n \sim q_{\theta_k}(\tau) \) for \( n \in 1:N \)\; \Comment{Sample proposal trajectories}

            \( \mathcal{D} \leftarrow \mathcal{D} \cup \{f(\tau'_n), p(\tau'_n) / q_{\theta_k}(\tau'_n)\}_{n=1}^N \)\; \Comment{Update IS buffer}

            \( \tilde{w}'_n \leftarrow p(\tau'_n) P_{\beta}(f(\tau'_n) - \gamma) / q_{\theta_k}(\tau'_n) \) for \( n \in 1:N \)\; \Comment{Compute proposal MH weights}

            \( \tau_n, \tilde{w}_n \leftarrow \MHAccept(\tau'_n, \tilde{w}'_n, \tau_n, \tilde{w}_n) \) for \( n \in 1:N \)\; \Comment{Accept or reject proposal trajectories}

            \( \mathcal{L}(\theta) \leftarrow \sum_{n=1}^N \sum_{t=1}^T -\log q_{\theta}(x_{t, n} \mid s_{t, n}) \)\; \Comment{Compute loss using \cref{eq:expanded-loss}}

            \( \theta_{k+1} \leftarrow \textproc{SGD}(\mathcal{L}, \theta_k) \)\; \Comment{Update proposal using stochastic gradient descent}
        }

        \Return \( \hat{\mu} \leftarrow \Estimate(\mathcal{D}, \gamma) \)\; \Comment{Compute failure probability estimate using \cref{eq:is-estimate}}
    }
\end{algorithm}

%% file: figures/pendulum.tex
\tikzset{
    pics/carc/.style args={#1:#2:#3}{
        code={
            \draw[pic actions] (#1:#3) arc(#1:#2:#3);
        }
    }
}

\begin{tikzpicture} [thick]

\definecolor{myred}{RGB}{245,97,92}
\definecolor{steelblue}{rgb}{0.27,0.51,0.71}
\definecolor{pastelRed}{RGB}{245, 97, 92}
\definecolor{pastelBlue}{RGB}{0, 114, 178}
\definecolor{pastelGreen}{RGB}{0, 158, 115}
\definecolor{pastelPurple}{RGB}{135, 112, 254}

\newcommand{\pang}{-45}
\newcommand{\fang}{-0}


\begin{scope} [draw = black, thin,
	fill = pastelBlue, 
	dot/.style = {black, radius = .025}]

\filldraw[opacity=0.2] [rotate around = {-\pang:(0,1.5)}] (.09,1.5) -- 
	node [midway, right] {} 
	node [very near end, right] {}
	+(0,2) arc (0:180:.09) 
	coordinate [pos = .5] (T) -- (-.09,1.5);
	
\filldraw[opacity=0.2] [rotate around = {\pang:(0,1.5)}] (.09,1.5) -- 
	node [midway, right] {} 
	node [very near end, right] {}
	+(0,2) arc (0:180:.09) 
	coordinate [pos = .5] (T) -- (-.09,1.5);


\filldraw (-0.1,1.5) -- coordinate [pos = .5] (F)
	(-0.1,1.4) -- node [above = .3cm] {}
	(0.1,1.4) -- (0.1,1.4) 
	coordinate (X) -- (.1,1.5)
	arc (0:180:.1) -- (-.1014,1.5);

\fill [dot] (0,1.52) circle;
\end{scope}

\begin{scope} [thin, black]
	\draw[opacity=0.0] (T) -- (0,1.52) coordinate (P);
	\draw [dashed] (P) + (0,-.1) -- +(0,2.0);
	\draw (P) + (0,0.75) arc (90:90-\pang:.75) node [black, midway, above] {$\theta$};
\end{scope}


\draw[dashed, color=pastelRed,] (0,1.5) -- +(60:2cm);
\draw[dashed, color=pastelRed,] (0,1.5) -- +(120:2cm);

\node[color=myred] at (0.0, 3.8){\scriptsize Failure: $|\theta|>\frac{\pi}{4}$ rad};



\end{tikzpicture}

%% file: figures/crosswalk.tex
\begin{tikzpicture}[x=0.04cm,y=-0.04cm, dot/.style = {circle, fill, minimum size=#1,
              inner sep=0pt, outer sep=0pt},
dot/.default = 1.2pt  
                    ]

  \begin{scope}

    \clip [rounded corners=3] (0, -5) rectangle (60, 29);

    \fill [gray!50] (0,-5) rectangle (60,29);
    \fill [green!40!black!40] (0, -5) rectangle (60, 0); 
    \fill [green!40!black!40] (0, 16) rectangle (60, 29);  

    \draw [white, dashed, thick] (0,8) -- (60,8);  

    


    \node[sedan top, body color=pastelBlue, window color=black!80,minimum width=0.5cm,](sedan) at (8, 12){};
    \node[circle, fill=pastelRed, inner sep=0pt, minimum size=3pt](ped) at (40, 22){};
    \node[](sedanarrow) at (25, 12){};
    \node[](sedanstart) at (11, 12){};
    \draw [->, >=stealth] (sedanstart) -- (sedanarrow);

    \node[](pedarrow) at (40, 13){};
    \draw [->, >=stealth] (ped) edge (pedarrow);

  \end{scope}

\end{tikzpicture}

%% file: figures/acas.tex
\def \globalscale {0.250000}
\begin{tikzpicture}[]

\node (ownship) [aircraft side,fill=pastelBlue,draw=pastelBlue,minimum width=1cm,scale=0.5] at (0,0) {};

\node (intruder) [aircraft side,fill=pastelRed,xscale=-1,draw=pastelRed,minimum width=1cm,scale=0.5] at (1.5, 0.2) {};
\draw [pastelBlue, dotted, line width=0.25mm, ->, >=stealth] (0.2,0) to[out=0,in=165] (1, -0.2);
\draw [pastelRed, dotted, line width=0.25mm, ->, >=stealth] (1.25, 0.2) to[out=-170,in=10] (0.5, 0.2);

\end{tikzpicture}

%% file: figures/f16.tex
\def \globalscale {0.250000}
\begin{tikzpicture}[]

\begin{scope}[xshift=15cm, rotate=-35, y=1cm, x=1cm, yscale=\globalscale,xscale=\globalscale, every node/.append style={scale=\globalscale}, inner sep=0pt, outer sep=0pt]
\path[fill] (10.3, 24.5) -- (11.9, 24.5) -- (12.1, 24.2) -- (15.0, 21.6) -- (17.4, 20.9) -- (17.4, 21.2) -- (17.5, 21.2) -- (17.8, 20.8) -- (20.1, 20.8) -- (22.1, 20.9).. controls (22.1, 20.9) and (23.5, 21.4) .. (24.5, 21.4).. controls (25.5, 21.4) and (26.3, 21.0) .. (27.3, 20.5).. controls (27.3, 20.5) and (28.2, 20.4) .. (28.3, 20.4).. controls (28.5, 20.4) and (30.2, 20.0) .. (30.8, 19.6) -- (31.0, 19.6) -- (31.0, 19.6) -- (31.5, 19.6) -- (31.5, 19.6) -- (31.6, 19.6) -- (31.6, 19.5) -- (32.2, 19.5) -- (32.2, 19.5) -- (31.7, 19.5) -- (31.7, 19.5) -- (31.5, 19.5) -- (31.5, 19.5) -- (31.0, 19.5) -- (31.0, 19.4) -- (30.8, 19.4) -- (30.8, 19.4).. controls (30.8, 19.4) and (29.7, 19.2) .. (28.3, 19.2).. controls (26.9, 19.2) and (25.0, 19.3) .. (24.9, 19.3) -- (24.9, 19.2).. controls (24.9, 19.2) and (24.7, 19.2) .. (24.7, 18.9) -- (24.7, 18.5) -- (24.2, 18.5) -- (24.1, 18.5).. controls (24.1, 18.5) and (18.1, 18.5) .. (17.9, 18.5).. controls (17.7, 18.5) and (16.5, 18.6) .. (16.4, 18.6).. controls (16.3, 18.5) and (16.0, 17.9) .. (16.0, 17.9) -- (14.3, 18.2) -- (14.3, 18.7).. controls (14.3, 18.7) and (12.6, 18.8) .. (12.0, 19.0).. controls (11.7, 19.0) and (10.9, 19.1) .. (10.8, 19.2) -- (10.8, 20.3).. controls (10.8, 20.3) and (11.2, 20.6) .. (11.8, 20.6) -- (12.2, 20.6) -- (12.2, 20.6) -- (11.6, 20.6) -- (11.6, 21.0) -- (11.4, 21.0) -- (11.4, 21.4) -- (11.6, 21.4) -- (10.3, 24.1).. controls (10.3, 24.1) and (10.2, 24.3) .. (10.2, 24.3).. controls (10.2, 24.4) and (10.2, 24.5) .. (10.3, 24.5) -- cycle(12.3, 20.6) -- (12.3, 20.5).. controls (14.5, 20.7) and (16.4, 20.7) .. (17.6, 20.7) -- (17.6, 20.8).. controls (16.0, 20.8) and (13.9, 20.7) .. (12.3, 20.6) -- cycle(21.8, 19.3) -- (24.8, 19.3) -- (24.8, 19.3) -- (21.8, 19.3) -- (21.8, 19.3) -- cycle(20.8, 19.2) -- (20.8, 19.5).. controls (21.9, 19.6) and (27.3, 19.5) .. (28.2, 19.4) -- (28.2, 19.5).. controls (27.3, 19.6) and (21.8, 19.7) .. (20.7, 19.6) -- (20.7, 19.6) -- (20.7, 19.3) -- (15.7, 19.3) -- (15.7, 19.5) -- (15.6, 19.5) -- (15.6, 20.0) -- (13.8, 20.0) -- (13.8, 19.8) -- (13.9, 19.8) -- (13.9, 19.9) -- (15.5, 19.9) -- (15.5, 19.4) -- (15.6, 19.4) -- (15.6, 19.2) -- cycle(14.2, 18.7) -- (14.3, 18.7) -- (14.3, 18.8) -- (16.4, 18.8) -- (16.4, 18.7) -- (16.5, 18.7) -- (16.5, 18.8) -- (14.2, 18.8) -- (14.2, 18.7) -- cycle(11.9, 19.9) -- (10.9, 19.7) -- (10.9, 19.3) -- (12.2, 19.2) -- (12.2, 19.2) -- (14.1, 19.7) -- (14.1, 19.8) -- (12.2, 19.3) -- (11.0, 19.4) -- (11.0, 19.7) -- (11.9, 19.8) -- cycle(12.1, 19.8) -- (12.1, 19.8) -- (13.8, 19.8) -- (13.8, 19.8) -- cycle;
\end{scope}

\fill[green!40!black!40] (15, -20) rectangle (50, -15);


\draw[pastelBlue, line width=2.5mm] (25, -1) -- ++(-35:10);


\draw [pastelBlue, dotted, line width=2mm, ->, >=stealth] (25, -1) ++(-35:10) to[out=-35,in=-120] (47, -6);

\draw [pastelRed, dotted, line width=2mm, ->, >=stealth] (25, -1) ++(-35:10) to (47, -15);

\end{tikzpicture}